\documentclass[acmsmall]{acmart}

\usepackage{algorithm}
\usepackage{algorithmic}

\usepackage{amsmath}
\usepackage{booktabs}
\usepackage{multirow}
\usepackage{threeparttable}
\usepackage{adjustbox}
\usepackage{pifont}
\usepackage[switch]{lineno}

\usepackage{subfigure}

\usepackage{epsfig}
\usepackage{rotating}
\usepackage{bbding}
\usepackage{enumitem}
\usepackage{bm}
\usepackage{colortbl}

\definecolor{mygray}{gray}{0.6}
\definecolor{mygray-bg}{gray}{0.9}
\definecolor{mypink}{cmyk}{0, 0.7808, 0.4429, 0.1412}
\definecolor{mygreen}{rgb}{0.0, 0.7, 0.0}
\definecolor{myblue}{rgb}{0.0, 0.72, 0.92}

\newcommand{\etal}{{\textit{et al}. }}
\newcommand{\ie}{{\textit{i}.\textit{e}.}}
\newcommand{\eg}{{\textit{e}.\textit{g}.}}
\newcommand{\etc}{\textit{etc.}}

\AtBeginDocument{%
  }

\setcopyright{acmcopyright}
\copyrightyear{2023}
\acmYear{2023}
\acmDOI{XXXXXXX.XXXXXXX}

\acmISBN{978-1-4503-XXXX-X/18/06}

\begin{document}

\title{Zero-Shot Scene Graph Generation via Triplet Calibration and Reduction}

\author{Jiankai Li}
\email{lijiankai@buaa.edu.cn}
\author{Yunhong Wang}
\email{yhwang@buaa.edu.cn}
\author{Weixin Li}
\email{weixinli@buaa.edu.cn}
\authornote{Corresponding author.}
\affiliation{%
  \institution{State Key Laboratory of Software Development Environment, School of Computer Science and Engineering, Beihang University; and Shanghai Artificial Intelligence Laboratory}
  \country{China}
  \postcode{100191}
}



\begin{abstract}
Scene Graph Generation (SGG) plays a pivotal role in downstream vision-language tasks.
Existing SGG methods typically suffer from poor compositional generalizations on unseen triplets.
They are generally trained on incompletely annotated scene graphs that contain dominant triplets and tend to bias toward these seen triplets during inference.
To address this issue, we propose a Triplet Calibration and Reduction (T-CAR) framework in this paper. In our framework, a triplet calibration loss is first presented to regularize the representations of diverse triplets and to simultaneously excavate the unseen triplets in incompletely annotated training scene graphs.
Moreover, the unseen space of scene graphs is usually several times larger than the seen space since it contains a huge number of unrealistic compositions. 
Thus, we propose an unseen space reduction loss to shift the attention of excavation to reasonable unseen compositions to facilitate the model training.
Finally, we propose a contextual encoder to improve the compositional generalizations of unseen triplets by explicitly modeling the relative spatial relations between subjects and objects.
Extensive experiments show that our approach achieves consistent improvements for zero-shot SGG over state-of-the-art methods. The code is available at \url{https://github.com/jkli1998/T-CAR}.
\end{abstract}




\maketitle

\section{Introduction}
Scene Graph Generation (SGG), which aims to detect object instances and their pairwise visual relationships, is crucial to visual comprehension \cite{krishna2017visual}. 
Such objects and visual relationships provide a compact and structured description of scenes, which can be used for high-level Vision-Language tasks, \eg \ visual question answering \cite{shi2019explainable,ben2019block,Li2022Inner,Liu2022Answer}, image captioning \cite{chen2020say,zhong2020comprehensive,wu2022learning,Yuan2020Image}, image retrieval \cite{dhamo2020semantic,guo2020visual,Yu2022Domain,Yanagi2022Intera}, \etc

\begin{figure}[!t]
    \centering
    \includegraphics[width = 0.98 \columnwidth]{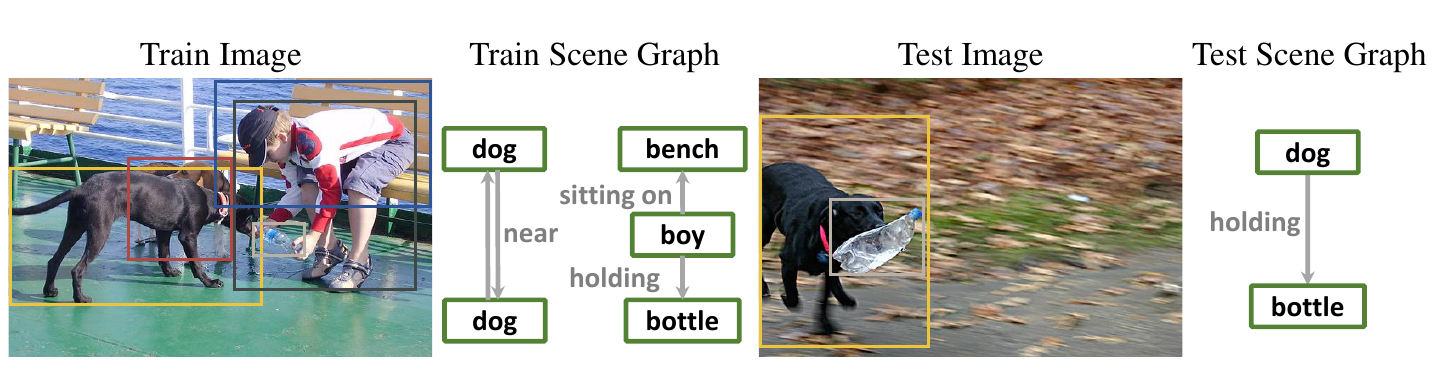}
    \caption{
Zero-shot scene graph generation aims to learn from triplets \textless subject-predicate-object\textgreater \ in the training set and infer their unseen compositions at test time. 
    }
    \label{fig:intro1}
\end{figure}

In the literature, SGG is typically formulated as predicting a triplet tuple \textless subject-predicate-object\textgreater \ \cite{xu2017scene,teng2022structured}. 
As shown in Fig. \ref{fig:intro1}, methods for zero-shot SGG need to learn from training triplets and then infer unseen triplets from test images. 
It is easy for us humans to recognize \textless dog-holding-bottle\textgreater \ from \textless boy-holding-bottle\textgreater \ and \textless dog-near-dog\textgreater \ since we acknowledge the concepts of ``boy'', ``dog'', ``bottle'', and ``holding''. 
But it is extremely challenging for SGG models when facing the unseen composition \textless dog-holding-bottle\textgreater \ \cite{tang2020unbiased,knyazev2021generative} because they suffer from the problem of biased seen triplets and are insensitive to unseen compositions.
Since less common triplets typically contain more information (according to the information entropy), this failure greatly impacts downstream tasks.

To improve the compositional generalization ability, previous works employ causal inference for unbiased prediction \cite{tang2020unbiased} or devise a Generative Adversarial Network (GAN) to generate unseen triplets \cite{knyazev2021generative}. Despite their remarkable progress in SGG, their performances in the zero-shot recall are still far from satisfactory. We attribute this weakened compositional generalization ability to the seen triplet bias, which mainly stems from two aspects. 

\begin{figure}[!t]
    \centering
    \includegraphics[width = 0.75 \columnwidth]{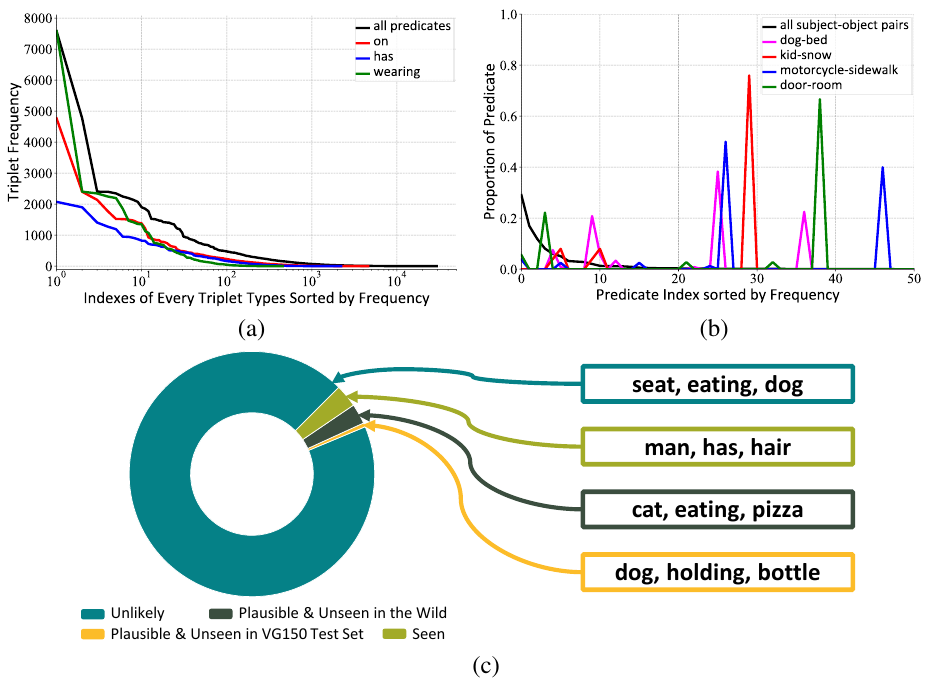}
    \caption{
(a) Several triplets dominate the Visual Genome \cite{krishna2017visual} benchmark, and (b) the distribution of predicates may be opposite to the overall distribution of the dataset with the same subject and object. 
(c) Unseen triplet space in Visual Genome is very large, but most of them are unlikely to be present.
    }
    \label{fig:intro2}
\end{figure}

On the one hand, dominant triplets lead to poorly discriminative representations of diverse triplets and bias the SGG model toward predicting frequent triplets. As shown in Figs. \ref{fig:intro2} (a) and \ref{fig:intro2} (b), several triplets dominate the dataset, even in the top three frequent predicates (\ie ``on'', ``has'', and ``wearing''). Moreover, given different pairs of subjects and objects, distributions of triplets are also dominated by several predicates, which may coincide with the distribution of predicates in the whole dataset or the opposite. 
On the other hand, recent works point out that the large-scale SGG benchmark contains many unlabeled, relatively rare, and meaningful relationships \cite{li2022devil,goel2022not}. These unseen triplets in the incompletely annotated training set are suppressed by classical cross-entropy loss, which forces the SGG model to bias toward predicting seen triplets.
Previous works focus on the predicate-granularity re-balance and unseen sample mining rather than the triplet-granularity \cite{tang2020unbiased,goel2022not,li2022devil}, resulting in unsatisfactory compositional generalization ability to zero-shot scene graph generation.
We consider a calibration method to regularize the discrimination of diverse triplets and excavate unseen triplets to address the above two issues.
However, exploring reasonable unseen triplets in an enormous unseen space is difficult. As shown in Fig. \ref{fig:intro2} (c), the unseen space is almost 37 times larger than the seen space in the Visual Genome dataset \cite{krishna2017visual}. Thanks to the realistic meaning of the scene graph, most of the compositions of unseen triplets are unlikely to be present in the real world, \eg \ \textless seat-eating-dog\textgreater \ \cite{knyazev2021generative}. Thus we further consider reducing the triplet space to shift the attention of our model to reasonable unseen compositions.

Based on the aforementioned observations, we propose a Triplet Calibration and Reduction (T-CAR) framework for zero-shot SGG to improve the model's generalization ability to unseen triplets. 
To this end,  we first introduce a Triplet Calibration Loss (TCL) to regularize the discriminative representations of diverse triplets and excavate unseen triplets.
TCL assigns triplet-specific calibrations on seen triplets to mitigate the bias toward frequent triplets and excavates unseen triplets to resist their negative constraints imposed by cross-entropy.
We then devise an Unseen Space Reduce Loss (USRL) to reduce the hindrance of mining unlabeled samples in such a huge unseen space with a large number of unreasonable compositions.
USRL exploits the interchangeability of subjects, predicates, and objects to explore the rationality of unseen compositions based on seen triplet samples, which is reformulated as the positive-unlabeled learning (PU Learning) problem. It regards all the seen triplets in the training set as in-positive data and all background triplets as unlabeled data. 
Finally, a Contextual Encoding Network (CEN) 
is further proposed to encode the spatial relationships between subjects and objects. Compared with previous context models \cite{zellers2018neural,tang2019learning,li2021bipartite}, it removes the linguistic priors and strengthens the relative position knowledge to reduce the seen triplet bias. 

Extensive experiments on the Visual Genome dataset \cite{krishna2017visual} are conducted to verify the effectiveness of our proposed modules. The experimental results demonstrate that our method outperforms state-of-the-art methods by a significant margin, \ie, \textbf{12.5\%}, \textbf{4.4\%}, and \textbf{1.8\%} of Zero-Shot Recall (zR@100) respectively for PredCls, SGCls, and SGDet tasks.

The main contributions of our work are summarized as follows: 

(1) A triplet calibration loss is introduced to balance the constraints on unseen triplets that are incorrectly annotated as background and enhance the attention of the model on less frequent triplet types to improve the model capability in representing diverse triplet compositions.

(2) An unseen space reduction loss is proposed to reduce the huge unseen triplet space. The loss allows the model to search for reasonable unseen compositions in a small triplet space restricted by linguistic priors. 

(3) A contextual encoding network is devised to explicitly encode the relative spatial features into triplet representations. Experiment results show that the relative spatial features are beneficial to distinguish unseen triplets.

\section{Related Work}
\subsection{Scene Graph Generation}
Scene graph is a structured representation of image content, bridging vision and language. It is involved in and facilitates many visual-language tasks \cite{gu2019unpaired,dhamo2020semantic}.

Some early works on scene graph generation explore incorporating more knowledge from various modalities \cite{lu2016visual,liang2018visual,zareian2020learning,zhong2021learning,sharifzadeh2022improving,Zhang2022Boosting}, and other approaches study the context modeling of entities and relations \cite{wang2020sketching,lu2021context,lin2022hl,chen2022resistance,lyu2022fine,dong2022stacked}.
Besides, scene-parsing-based models \cite{songchun2022cascaded,chen2019holistic++,wang2021hierarchical} also demonstrate strong abilities in parsing relationships, \eg, the cascaded scene parsing model which performs relation reasoning stage by stage, and achieves promising results \cite{songchun2022cascaded}. A recently published survey \cite{chang2023survey} conducts a comprehensive investigation of current scene graph researches, showing the development of scene graph generation models.
Although much progress has been made in the last few years in generating scene graphs on seen samples, existing methods still perform poorly in generating unseen triplets. The challenge of correctly predicting unseen triplets is not the same as the unbalanced predicate problem since frequent predicates and objects in the test set also dominate unseen triplets \cite{knyazev2021generative}. 
Various methods are proposed to address the zero-shot generalization problem.
Some approaches try to increase the diversity of input scene graphs by augmenting scene graphs based on Generative Adversarial Network (GAN) \cite{knyazev2021generative}. 
Some methods mitigate training bias through the model designed, where they either directly generate triplets instead of following the two-stage paradigm \cite{teng2022structured} or draw the counterfactual causality \cite{tang2020unbiased}. Other approaches devise graph-normalized \cite{knyazev2020graph} or energy-based loss \cite{suhail2021energy} to improve the compositional generalization.
However, previous SGG works overlook the issues of large unseen triplet space containing lots of unrealistic triplets and unseen samples in the incompletely annotated training set. Their performance on zero-shot generalization is still far from satisfactory. 
Our work considers triplets calibration and unseen space reduction in a single framework. 
We propose a triplet calibration loss to regularize the triplet representations and excavate unseen triplets, and an unseen space reduction loss to reduce unseen space and shift the attention to reasonable unseen triplets.

\subsection{Compositional Zero-Shot Learning}
Compositional zero-shot learning (CZSL) is subordinate to zero-shot learning \cite{Xu2021Zero,pourpanah2022review,Rao2022Dual}, which aims to transfer knowledge from seen to unseen compositions. And the goal of zero-shot SGG is to transfer knowledge from seen combinations of subjects, predicates, and objects to unseen combinations. Thus zero-shot SGG can be regarded as CZSL.

Typical CZSL works \cite{huynh2020compositional,mancini2021open,Karthik_2022_CVPR,li2022siamese} focus on the combinations of objects and states, which is different from zero-shot SGG. Compared to these works, zero-shot SGG owns a larger label space since it has three degrees of freedom.
Taking the MIT-States \cite{isola2015discovering}, one of the biggest and most commonly used datasets for CZSL, as an example, there are 1,262 seen and 400 unseen compositions in the training and test sets, respectively. 
In contrast, VG150 \cite{krishna2017visual} is composed of about 29 thousand seen, 5 thousand unseen, and 1 million potentially existing compositions in the training and test sets. 
In this paper, our T-CAR narrows down the huge potential triplet space, removes the most unconventional combinations, and reduces the learning and inference difficulties.

\section{Our Approach}

\subsection{Problem Formulation and Overview}
\subsubsection{Problem Formulation.}
Given an image $\mathcal{I}$, the task of scene graph generation (SGG) is to predict a graph $\mathcal{G}=(\mathcal{V},\mathcal{E})$, where $\mathcal{V}$ is the entity node set and $\mathcal{E}$ denotes the set of relationships of two ordered entities. Each entity node $v \in \mathcal{V}$ is composed of a bounding box $\textit{\textbf{b}}$, a visual feature $\textit{\textbf{v}}$, and a class label $c_v \in \mathcal{C}_e$. A relationship $r=(s,p,o) \in \mathcal{E}$ is a three-tuple, including a subject entity node $s$, an object entity node $o$, and its predicate label $p \in \mathcal{C}_p$. 

Scene graph generation is typically a three-stage task, which can be formulated as follows:
\begin{align}
    P(\mathcal{G}|\mathcal{I}) = P(\mathcal{B}|\mathcal{I}) P(\mathcal{C}_{e}|\mathcal{B},\mathcal{I}) P(\mathcal{G}|\mathcal{C}_{e},\mathcal{B},\mathcal{I}),
\end{align}
where $P(\mathcal{B}|\mathcal{I})$ aims to detect bounding boxes of entities, $P(\mathcal{C}_{e}|\mathcal{B},\mathcal{I})$ works on predicting the entity categories, and $P(\mathcal{G}|\mathcal{C}_{e},\mathcal{B},\mathcal{I})$ denotes the predicate classification.

\begin{figure*}[!tb]
\centering {\includegraphics[width=1.0\textwidth]{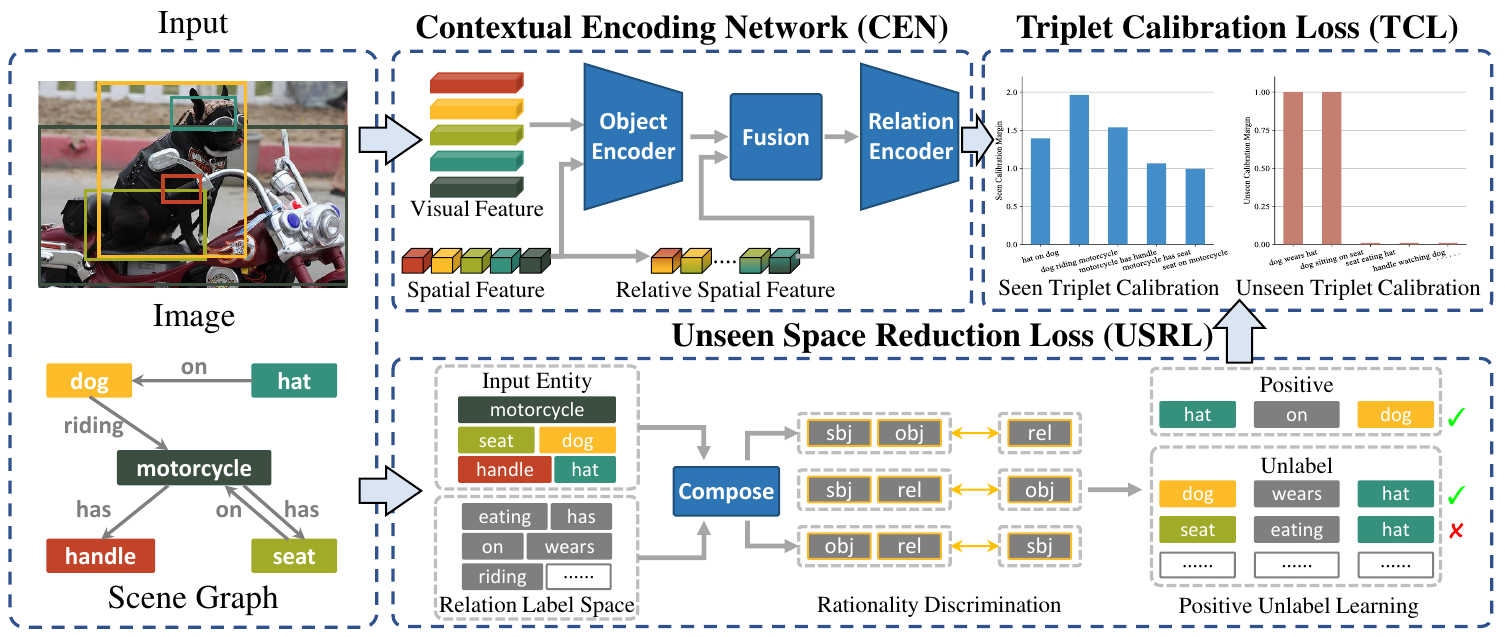}}
\caption{
The overall framework of our T-CAR model. Given an input image, T-CAR first detects its object proposals. Then the CEN encodes the entity and relation features. Meanwhile, USRL reduces a large number of impossible unseen triplets in the output space. 
TCL imposes calibration margins to each seen triplet based on its frequency, shifting attention to rare triplets and regularizing discriminative representations of diverse triplets. It also excavates unseen triplets in the training set according to the reasonable triplets provided by USRL.
}
\label{fig:framework}
\end{figure*}

\subsubsection{Method Overview.}
An overview of our Triplet Calibration and Reduction (T-CAR) model is illustrated in Fig. \ref{fig:framework}. 
A contextual encoding network (CEN) is first introduced to generate context-aware representations of entity nodes and relationships with less seen triplet bias. Compared to previous contextual encoders, CEN removes the linguistic prior and explicitly models the relative positions between subjects and objects to reduce seen triplet bias and improve compositional generalization ability.
After that, we propose a triplet calibrate loss (TCL) to alleviate the bias effect of dominant seen triplets and mine the unseen compositions. 
However, the unseen triplet space is very large and typically contains a huge number of unrealistic compositions. It is difficult to excavate and infer unseen samples in such an enormous space.
We observe the substitutability of semantically similar terms between seen triplets. Based on this observation, we propose the unseen space reduction loss (USRL) for training and inference, eliminating the most unrealistic triplets based on linguistic knowledge.

\subsection{Contextual Encoding Network}
\label{sec:context}
It is a long-standing paradigm in SGG to initialize entity representations with visual and linguistic (from their text labels) features and initialize relation representations with visual and spatial features of the subject and object. However, we find that the linguistic features bring a bias to the seen compositions, while the relative position cues between subject and object are neglected. From our perspective, these linguistic features, fixed for a certain class of entities, cause the SGG model to learn the conditional distribution from the composition priors of the subject and object \cite{zellers2018neural}. These composition priors contribute to seen triplets but weaken the ability to represent unseen compositions.
Moreover, relative spatial cues are also indispensable for robust relationship predictions. Most existing methods \cite{zellers2018neural,tang2019learning,li2021bipartite} in SGG simply concatenate or sum up the spatial features of subjects and objects, which may be insufficient to mine the underlying spatial relations.
To address these deficiencies, 
we propose a contextual encoding network, which removes the linguistic priors and explicitly models the relative spatial features. Our contextual encoding network is composed of an entity encoder, a fusion layer, and a relation encoder. 

The entity encoder is responsible for refining entity features by interacting with contextual entities:
\begin{equation}
\begin{aligned}
\{\textit{\textbf{x}}_i\}_{i=1,...,N} = Enc_{obj}(\{[\textit{\textbf{v}}_i, FFN(\textit{\textbf{b}}_i)]\}_{i=1,...,N}),
\end{aligned}
\end{equation}
where $\textit{\textbf{x}}_i$ denotes the refined entity representations, $N$ is the number of entities, $[\cdot,\cdot]$ represents the concatenation operation, $FFN(\cdot)$ is a two-layer Multi-Layer Perceptron (MLP) with LeakyReLU activation, and $Enc_{obj}$ denotes the entity contextual modules which could be the multi-layer LSTM \cite{hochreiter1997long}, GNN \cite{hamilton2017inductive}, or Transformer \cite{vaswani2017attention}. Here we apply a four-layer Transformer as the entity encoder. Note that the object features are initialized without linguistic features. 

The fusion layer aims to initialize predicate representation $\textit{\textbf{x}}_p'$ with refined entity representations of its corresponding subject $\textit{\textbf{x}}_s$ and object $\textit{\textbf{x}}_o$:
\begin{equation}
\begin{aligned}
\textit{\textbf{x}}'_p = \textit{\textbf{x}}_s \star \textit{\textbf{x}}_o \star FFN([\textit{\textbf{v}}_u, \textit{\textbf{b}}_{s,o}]),
\end{aligned}
\end{equation}
where $\textit{\textbf{v}}_u$ is extracted from the union box of subject $s$ and object $o$. The operation $\star$ denotes the fusion function defined in \cite{tang2019learning}: $\textit{\textbf{x}} \star \textit{\textbf{y}} = ReLU(\textit{\textbf{W}}_x\textit{\textbf{x}} + \textit{\textbf{W}}_y\textit{\textbf{y}})-(\textit{\textbf{W}}_x\textit{\textbf{x}} - \textit{\textbf{W}}_y\textit{\textbf{y}}) \odot (\textit{\textbf{W}}_x\textit{\textbf{x}} - \textit{\textbf{W}}_y\textit{\textbf{y}})$, where $\odot$ denotes the Hadamard product, and $\textit{\textbf{W}}_x$ and $\textit{\textbf{W}}_y$ are parameters used for projection. 

The relative spatial feature $\textit{\textbf{b}}_{s,o}$ is calculated from subject and object bounding boxes, \ie \ $\textit{\textbf{b}}_s = (x_{s}^1, y_{s}^1, x_{s}^2, y_{s}^2)$  and $\textit{\textbf{b}}_o = (x_{o}^1, y_{o}^1, x_{o}^2, y_{o}^2)$, where $(x_{\cdot}^1,y_{\cdot}^1)$ and $(x_{\cdot}^2,y_{\cdot}^2)$ are the coordinates of corner points of the bounding box. The relative spatial feature is composed of the normalized union box location $\textit{\textbf{b}}_{ul}$, relative size $\textit{\textbf{b}}_{sl}$, and relative location $\textit{\textbf{b}}_{rl}$ since they imply potential categories of the relationships between subjects and objects:

\begin{align}
\textit{\textbf{b}}_{ul} = &(\frac{x_{u}^1}{w}, \frac{y_{u}^1}{h}, \frac{x_{u}^2}{w}, \frac{y_{u}^2}{h}, 
                            \frac{x_{u}^1 + x_{u}^2}{2w}, \frac{y_{u}^1 + y_{u}^2}{2h}, \frac{w_u}{w}, \frac{h_u}{h}), \\
\textit{\textbf{b}}_{sl} = &({\rm log}(\frac{w_s}{w_o}), {\rm log}(\frac{h_s}{h_o}), {\rm log}(\frac{w_o}{w_s}), {\rm log}(\frac{h_o}{h_s})), \\
\textit{\textbf{b}}_{rl} = &(\frac{x_{s}^1-x_{o}^1}{w_o}, \frac{y_{s}^1-y_{o}^1}{h_o}, \frac{x_{s}^2-x_{o}^2}{w_o}, \frac{y_{s}^2-x_{o}^2}{h_o}, \nonumber \\
                         & \frac{x_{o}^1-x_{s}^1}{w_s}, \frac{y_{o}^1-y_{s}^1}{h_s}, \frac{x_{o}^2-x_{s}^2}{w_s}, \frac{y_{o}^2-y_{s}^2}{h_s}), 
\end{align}
where $(w, h)$, $(w_s, h_s)$, and $(w_o, h_o)$ denote the widths and heights of image, subject, and object, respectively. $(x_{u}^1, y_{u}^1)$ and $(x_{u}^2, y_{u}^2)$ are the coordinates of the corner points of the union bounding box of subject $\textit{\textbf{b}}_s$ and object $\textit{\textbf{b}}_o$. We concatenate these features and obtain the relative spatial feature $\textit{\textbf{b}}_{s,o}=[\textit{\textbf{b}}_{ul},\textit{\textbf{b}}_{sl},\textit{\textbf{b}}_{rl}]$.

The relation encoder aims to obtain the refined predicate feature $\textit{\textbf{z}}_j$:
\begin{equation}
\begin{aligned}
\{\textit{\textbf{z}}_j\}_{j=1,...,N\times(N-1)} = Enc_{rel}(\{\textit{\textbf{x}}_j'\}_{j=1,...,N\times(N-1)}),
\end{aligned}
\end{equation}
where $Enc_{rel}$ denotes the contextual module encoding the relation context. Here we apply a two-layer Transformer as the relation encoder. Finally, we decode entity class logits $\textit{\textbf{e}}_i$ and predicate class logits $\textit{\textbf{r}}_j$ through two fully-connected layers:
\begin{align}
\textit{\textbf{e}}_i = FC_{obj}(\textit{\textbf{x}}_i), \\
\textit{\textbf{r}}_j = FC_{rel}(\textit{\textbf{z}}_j),
\end{align}
where $FC_{obj}(\cdot)$ and $FC_{rel}(\cdot)$ denote the fully-connected layers.

Our CEN fuses the contextual feature with object features and models the relationships among entities, which shares a similar framework with the Object-centric Feature Alignment Module \cite{wang2021symbiotic}. However, the key contribution of CEN lies in the relative spatial relationship modeling. It explicitly computes the relative spatial features between subjects and objects, while directly taking the spatial feature, \ie \ bounding box, or implicitly modeling their spatial relationships through ROI features like OFAM \cite{wang2021symbiotic} is insufficient for generating complicated unseen triplet compositions.

\subsection{Triplet Calibration Loss}
\label{sec:calibrate}
Existing SGG methods \cite{zellers2018neural,tang2019learning,li2021bipartite} typically utilize cross-entropy loss on relations to optimize their SGG models. 
However, optimizing cross-entropy loss on scene graphs consisting of a number of dominant triplets and unseen triplets is prone to suppressing the probabilities of unseen triplets and making the model biased towards these dominant seen triplets. 
In other words, given the fixed subject and object entities, 
these models will predict high probabilities for high-frequent seen triplets, obstructing the compositional generalization. 

\subsubsection{Unseen Triplet Calibration.} To address these deficiencies, we first devise a calibration loss on unseen samples to resist cross-entropy constraints during training:
\begin{align}
\mathcal{L}_{cal}(\textit{\textbf{r}}_{s,o}) = - {\rm log} \, (\sum_{(s,c,o) \in \mathcal{C}^{u}_{tpt}} \frac{\exp(r_{s,o}^c)}{\sum_{c'\in \mathcal{C}_p \exp(r_{s,o}^{c'})}}),
\end{align}
where $\mathcal{C}^{u}_{tpt}$ denotes the unseen triplet set, $r^{c}_{s,o}$ is the logit  that the relationship category between the subject $s$ and the object $o$ belongs to class $c$, which is the $c$-th element in decoded $\textit{\textbf{r}}_{s,o}$. 
Minimization of $\mathcal{L}_{cal}$ promotes these unseen samples to have a non-zero probability during training and improves the generalization ability of the SGG model toward unseen triplets. 
But for those predicates that are correctly annotated, 
very small logits on unseen triplets are capable of incurring a huge loss with the amplification of the function $log(\cdot)$ and affecting the correct label, which is not desired.  
Here we apply a hard margin to reduce the effect of $\mathcal{L}_{cal}$ on the annotated category:
\begin{equation}
\begin{aligned}
    \mathcal{L}_{cal}^m(\textit{\textbf{r}}_{s,o}) 
    =& \mathcal{L}_{cal}(\textit{\textbf{r}}_{s,o}+ \mathbb{I}_{m}(s,:,o) ) \\
    =& - {\rm log} \, (\sum_{(s,c,o) \in \mathcal{C}^{u}_{tpt}} \frac{\exp(r^c_{s,o}+1)}
    {\sum_{c' \in \mathcal{C}_{tpt}^{s}} \exp(r^{c'}_{s,o} - 1) +
    \sum_{c''\in \mathcal{C}_{tpt}^{u}} \exp(r^{c''}_{s,o}+1)}),
    \label{eq:cal-m}
\end{aligned}
\end{equation}
where $\mathcal{C}_{tpt}^{s}$ denotes the seen triplets set, and $\mathbb{I}_{m}(s,:,o)\in \mathbb{R}^{|\mathcal{C}_p|}$ is a margin vector.
When $(s, c, o)$ is an unseen composition, the value of $\mathbb{I}_{m}(s,c,o)$ is $1$. Otherwise, it is $-1$. For those correctly labeled as background triplets, hard margins mitigate the problem of model training instability due to the large losses incurred by their small logits.
For those unseen triplets incorrectly labeled as background, the calibration loss works inversely with the cross-entropy loss, reducing the constraint on unseen samples and generating reasonable confidence to the unseen triplets during inference.

\subsubsection{Seen Triplet Calibration.}
The predicate distribution is diverse when given different subjects and objects, and it may coincide with the distribution of predicates in the whole dataset or maybe the opposite. Previous works \cite{zellers2018neural,dong2022stacked} consider the predicate distribution from the coarse granularity of the dataset rather than from the fine granularity of the subject-object combination, which weakens the ability of the model to represent various triplets. We propose a fine-grained method for calibrating seen samples that collaborates with unseen calibration mitigating the bias towards frequently seen triplets while enhancing the attention on rare and diverse compositions. We consider adjusting the margins of seen triplets:
\begin{equation}
\begin{aligned}
    \mathcal{L}_{cal}^{m, \alpha}(\textit{\textbf{r}}_{s,o})
    =& \mathcal{L}_{cal}(\textit{\textbf{r}}_{s,o}+ \mathbb{I}_{m}^{\alpha}(s,:,o) ) \\
    =& - {\rm log} \, (\sum_{(s,c,o) \in \mathcal{C}^{u}_{tpt}}
    \frac{\exp(r^c_{s,o}+1)}
    {\sum_{c' \in \mathcal{C}_{tpt}^{s}}\exp(r^{c'}_{s,o}-\alpha_{s,c',o}) 
    + \sum_{c''\in \mathcal{C}_{tpt}^{u}}\exp(r^{c''}_{s,o}+1)}) ,
    \label{eq:loss_cal_alpha}
\end{aligned}
\end{equation}
where $\mathcal{L}_{cal}^{m, \alpha}(\textit{\textbf{r}}_{s,o})$ replaces the fixed seen triplet margin in $\mathcal{L}_{cal}^m(\textit{\textbf{r}}_{s,o})$ with a dynamic margin $\alpha$ conditioning on different seen triplets.
In Eq. \ref{eq:loss_cal_alpha}, the dynamic margin becomes the coefficient $\exp(-\alpha_{s,c',o})$ of its corresponding term $\exp(r_{s,o}^{c'})$ of the seen triplet. A smaller $\alpha_{s,c',o}$ will impose a larger constraint on its corresponding triplet during optimization. We expect frequent triplets to be subject to large constraints and rare compositions to be subject to relatively small constraints to balance the attention to various compositions. Thus, seen triplet calibration is designed to provide a small $\alpha$ for dominant triplets and a large value for rare triplets:
\begin{align}
     \alpha_{s,c,o}= {\rm log} \,(\frac{n_{max}}{n_{s,c,o}}) \times \frac{\sum_{i\in\mathcal{C}_{tpt}}n_{i}}{
     \sum_{j\in\mathcal{C}_{tpt}}n_{j} {\rm log} \,(\frac{n_{max}}{n_{j}})} ,
     \label{eq:alpha}
\end{align}
where $n_{max}$ and $n_{s,c,o}$ are the counts of triplet with the largest number and triplet $(s,c,o)$, respectively. The first term in Eq. \ref{eq:alpha} is the margin weight, and the second term is used to normalize these margins. Margin $\alpha_{s,c,o}$ decreases as the count of $(s,c,o)$ increases, which in turn adjusts constraints for the different counts of triplets. In addition, we also add this margin to the cross-entropy loss in seen triplet calibration to reduce the bias of dominant seen triplets:
\begin{align}
    \mathcal{L}_{ce}^{m, \alpha}(\textit{\textbf{r}}_{s,o})
    = - {\rm log} \, \frac{\exp(r^c_{s,o}-\alpha_{s,c,o})}{\sum_{c'}\exp(r^{c'}_{s,o}-\alpha_{s,c',o})}.
\end{align}
Combining the unseen and seen triplet calibrations, the final loss of our method is:
\begin{align}
\label{eq:all-loss}
    \mathcal{L} = \mathcal{L}^{m, \alpha}_{ce} + \lambda \mathcal{L}_{cal}^{m,\alpha},
\end{align}
where $\lambda$ is a pre-defined weighting hyper-parameter.
During inference, we still calibrate the seen and unseen triplets as:
\begin{align}
    predicate = \mathop{\rm argmax}\limits_{c\in \mathcal{C}_p} \, r^{c}_{s,o} + \mathbb{I}_{m}(s,c,o).
    \label{eq:cal-infer}
\end{align}
So in the prediction, the model can regularize the discriminative representations of diverse triplets and well-consider unseen samples.

\subsection{Unseen Space Reduction}
\label{sec:usrl}
Almost all of the existing SGG approaches treat the entire triplet space as the space of their potential generation results. They assume that all unseen compositions are potentially possible triplets that could exist in the real world and will output a certain probability to these triplets. However, compared to the count of seen triplet categories, the space of the whole composition is very large (which is 1,125,000 vs 29,283 for VG150).
Only a small fraction of the unseen triplets could exist in the real world \cite{knyazev2021generative}. Moreover, most triplets are unrealistic compositions, \eg \ \textless seat-eating-dog\textgreater. Hence, reducing the unseen triplet space is necessary to enhance the mining results of unseen triplets during training and inference.

It is observed that there is interchangeability between seen triplets, where subjects, predicates, and objects with similar properties can be replaced to form new compositions (\eg \ \textless dog/elephant-walking on-street\textgreater, \textless man-using/holding-phone\textgreater, and \textless man-riding-house/elephant\textgreater). 
We propose an Unseen Space Reduction Loss (USRL) to reduce the unseen triplet space by starting from this interchangeability among subjects, predicates, and objects. 
We first fuse any two elements of the triplet and then project them onto the same space as another element to explore the rationality of this triplet:
\begin{align}
\left\{
\begin{array}{cc}
     \textit{\textbf{d}}_{s} = \sigma(\textit{\textbf{t}}_{p} \star \textit{\textbf{t}}_{o}) \odot
                                \sigma(\textit{\textbf{w}}_{s}\textit{\textbf{t}}_{s}) \\
     \textit{\textbf{d}}_{p} = \sigma(\textit{\textbf{t}}_{s} \star \textit{\textbf{t}}_{o}) \odot
                                \sigma(\textit{\textbf{w}}_{p}\textit{\textbf{t}}_{p})  \\
     \textit{\textbf{d}}_{o} = \sigma(\textit{\textbf{t}}_{s} \star \textit{\textbf{t}}_{p}) \odot
                                \sigma(\textit{\textbf{w}}_{o}\textit{\textbf{t}}_{o})
\end{array}
\right. ,
\end{align}
where $\textit{\textbf{t}}_{s}$, $\textit{\textbf{t}}_{p}$, and $\textit{\textbf{t}}_{o}$ denote the linguistic embeddings for subject, predicate, and object, respectively. $\textit{\textbf{w}}_{s}$, $\textit{\textbf{w}}_{p}$, and $\textit{\textbf{w}}_{o}$ are trainable parameters, and $\sigma$ represents the sigmoid function.
Then we judge the reasonableness of the triplet $d_{usrl}$ based on the results of the above three aspects:
\begin{align}
    d_{usrl} = \textit{\textbf{w}}_{usrl} [ \textit{\textbf{d}}_{s}, \textit{\textbf{d}}_{p}, \textit{\textbf{d}}_{o} ],
\end{align}
where $\textit{\textbf{w}}_{usrl}$ denote parameters that project the concatenated feature to one dimension. The process of learning knowledge from seen triplets and determining the rationality of unseen triplets can be formulated as the Positive-Unlabeled Learning problem \cite{bekker2020learning,zhao2022dist}. Specifically, based on the annotated compositions that are known to be positive, we classify which ones are negative from a bunch of unlabeled triplets. Here we apply the nnPU \cite{kiryo2017positive} to train the unseen space reduction:
\begin{equation}
\begin{aligned}
\label{eq:pu-loss}
    \mathcal{L}_{usrl}=& \frac{\pi}{n_{pos}} \sum_{h_i}\mathcal{L}_{bce}^{+}(h_i) + 
    \max(0,  
    & \frac{1}{n_{u}} \sum_{h_j}\mathcal{L}_{bce}^{-}(h_j) - \frac{\pi}{n_{pos}} \sum_{h_i}\mathcal{L}_{bce}^{-}(h_i)),
\end{aligned}
\end{equation}
where $\mathcal{L}_{bce}^{+}$ and $\mathcal{L}_{bce}^{-}$ denote the binary cross-entropy functions to calculate the risk of misclassifying the input into negative and positive samples, respectively. $n_{pos}$ and $n_u$ are the counts of positive and unlabeled samples, $\pi$ represent the fraction of positive samples, and $h_i$ and $h_j$ are the predicted confidences of positive sample and unlabeled sample, respectively. The negative compositions judged by the USRL will narrow the range of $\mathcal{C}_{tpt}^{u}$, which affects the margin vectors $\mathbb{I}_{m}^\alpha(s,c,o)$ and $\mathbb{I}_{m}(s,c,o)$ in Eq. \ref{eq:loss_cal_alpha} and Eq. \ref{eq:cal-infer}, respectively.

\section{Experiments}
\subsection{Experiment Setting}
\subsubsection{Dataset.} 
Our experiments for scene graph generation are conducted on the Visual Genome dataset \cite{krishna2017visual}.  We follow the most widely used VG150 split \cite{xu2017scene,zellers2018neural,tang2019learning,knyazev2021generative}, which contains the most frequent 150 object categories and 50 relation categories in Visual Genome.

\subsubsection{Tasks.}
We adopt the following three conventional evaluation tasks. 1) Predicate Classification (\textbf{PredCls}) aims to predict the predicates of pairwise relationships with ground-truth object bounding boxes and their object categories. 2) Scene Graph Classification (\textbf{SGCls}) aims to predict the predicates and object categories with ground-truth object bounding boxes. 3) Scene Graph Detection (\textbf{SGDet}) aims to detect object bounding boxes and categories in the image and predict their pairwise relationships.
\subsubsection{Evaluation Metric and Protocol.}

We evaluate SGG methods with image-wise recall evaluation metrics, including Recall@K (R@K) and Zero-Shot Recall@K (zR@K), which are also adopted in previous works \cite{suhail2021energy,knyazev2021generative,teng2022structured,goel2022not}. 
The R@K measures the fraction of ground-truth relationship triplets that appear in the top K most confident triplet predictions in an image, zR@K metric measures the fraction of zero-shot ground-truth relationship triplets that appear in the top K most confident triplet predictions in an image. We average these fractions across images to obtain R@K and zR@K separately. 
Note that we do not consider test images that do not contain zero-shot triplets for zR@K following \cite{tang2020unbiased,knyazev2020graph,suhail2021energy,knyazev2021generative,hung2021contextual,teng2022structured,goel2022not}. 
Previous test protocols for zero-shot scene graph generation are diverse \cite{tang2020unbiased,knyazev2020graph,suhail2021energy,knyazev2021generative,hung2021contextual,teng2022structured,goel2022not}, and some of them apply Frequency Bias \cite{zellers2018neural} that greatly degrades zero-shot performance (4.8 versus 20.5 for Motifs \cite{zellers2018neural} under zR@100 and Predcls). Our method cannot be directly compared with all previous approaches. Thus, we unify the test protocols and re-implement previous methods based on their source codes without Frequency Bias for fair comparisons. 

We first review the previous test protocols before introducing our unified test protocol. There are several key differences between previous test protocols.

1) \textbf{Object Overlap.}
The requirement to have overlap between the training objects is proposed by Zellers \etal \cite{zellers2018neural}.
This means that only relationships of objects that overlap with other objects are allowed to be used as training data.
They think objects that do not overlap with other objects typically own low-quality Region of Interest (RoI).
This requirement limits the relationships in the training set, resulting in less training data and more unseen samples in the corresponding test set.

2) \textbf{Validation Set.}
The most widely used split in Visual Genome dataset \cite{krishna2017visual} is VG150 \cite{xu2017scene}. VG150 is composed of a training set, a validation set, and a test set. The current mainstream \cite{tang2020unbiased,suhail2021energy,hung2021contextual,teng2022structured,goel2022not} zero-shot Scene Graph Generation (SGG) test protocol does not use the validation set. However, some methods \cite{knyazev2020graph,knyazev2021generative} apply the validation set, and thus their zero-shot test set does not contain triplets in the validation set.

3) \textbf{Frequency Bias.}
Frequency Bias \cite{zellers2018neural} is a trick to improve the performance of SGG. However, as shown in the paper, it will significantly damage the performance of the SGG model on unseen triplets.

There mainly exist three evaluation protocols used in previous works.

1) The first zero-shot SGG evaluation protocol\footnote{https://github.com/bknyaz/sgg} is built on top of \textit{neural-motifs}\footnote{https://github.com/rowanz/neural-motifs}. 
It does not require the object overlap, applies the validation set, and does not use the Frequency Bias. 
Methods, \eg \ \cite{knyazev2020graph,knyazev2021generative}, using this evaluation protocol apply VGG16 \cite{simonyan2014very} as their backbone. 

2) The second zero-shot SGG evaluation protocol\footnote{https://github.com/KaihuaTang/Scene-Graph-Benchmark.pytorch \label{ftn:kaihua}} is introduced by Tang \etal \cite{tang2020unbiased}. It requires the object overlap, does not applies the validation set, and uses the Frequency Bias. Methods \eg \ \cite{tang2020unbiased} using the second evaluation protocol apply ResNeXt-101-FPN \cite{lin2017feature} as their backbone.

3) The third zero-shot SGG evaluation protocol is built on top of \textit{Scene-Graph-Benchmark.pytorch}\textsuperscript{\ref{ftn:kaihua}}. 
It does not require the object overlap, does not apply the validation set, and uses the Frequency Bias. Methods \eg \ \cite{suhail2021energy,teng2022structured,goel2022not} using the third evaluation protocol apply ResNeXt-101-FPN \cite{lin2017feature} as their backbone.

Our unified zero-shot SGG evaluation protocol is also built on top of \textit{Scene-Graph-Benchmark.pytorch}. To exploit the whole training data, our evaluation protocol does not require object overlap, which is also the option used in most zero-shot SGG methods. It does not apply the validation set and removes the Frequency Bias to improve the performances on unseen triplets.

For comparison with VGG-16, we apply the same evaluation protocol as the comparison methods, and their results are obtained directly from the original papers.
For the experiments with ResNeXt-101-FPN, we apply our evaluation protocol and re-implement the comparison methods including IMP \cite{xu2017scene}, VTransE \cite{zhang2017visual}, Motifs \cite{zellers2018neural}, IMP++ \cite{knyazev2020graph}, TDE \cite{tang2020unbiased}, UVTransE \cite{hung2021contextual}, BGNN \cite{li2021bipartite}, EBM \cite{suhail2021energy}, GRAPHN \cite{knyazev2021generative}, and SSR \cite{teng2022structured} without Frequency Bias according to their source codes for fair comparisons.

\subsubsection{Implementation Details}
For fair comparison on VG, we adopt the pre-trained VGG-16 \cite{simonyan2014very} and ResNeXt-101-FPN \cite{lin2017feature} as the backbones. We follow previous methods and use GloVe \cite{pennington2014glove} with 200 dimensions as the linguistic embedding. We also apply the same post-processing method as previous methods \cite{lin2022hl,lin2022ru}, \ie \ relational-NMS, to filter the generated redundant triplets.
Our network is optimized by Stochastic Gradient Descent (SGD) with an initial learning rate of $10^{-3}$, and the batch size is set as 14. The number of total iterations is 16k, and the learning rate is decayed by the factor of 10 on the $10{\rm k}^{\rm th}$ iterations. The reduced prediction space accounts for 85\% of the total triplet space.
The parameters $\lambda$ and $\pi$ in Eq. \ref{eq:all-loss} and Eq. \ref{eq:pu-loss} are set as 0.01 and 0.03, respectively. Our codes are implemented with PyTorch and 2 NVIDIA GeForce RTX 2080Ti GPUs.

\begin{table}[tbp]
    \centering
    \caption{Performance comparison results of state-of-the-art SGG models on three SGG tasks with graph constraint. ``B" denotes the backbone of object detector (Faster R-CNN~\cite{ren2015faster}) used in each SGG model. \dag \, denotes that the results are obtained with an unknown evaluation protocol, and thus, may not be directly comparable. ``-'' represents that the result is not mentioned in the original paper or the method is unavailable in that configuration. 
    The \textcolor{red}{\textbf{best}} and \textcolor{myblue}{\textbf{second best}} results under each setting are marked in \textcolor{red}{\textbf{red}} and \textcolor{myblue}{\textbf{blue}}, respectively.
    }
        \resizebox{14cm}{!}{
            \begin{tabular}{c| r | c c c | c  c  c | c c c}
                \hline
                \multirow{2}{*}{B} &
                \multirow{2}{*}{Models} & \multicolumn{3}{c|}{PredCls} & \multicolumn{3}{c|}{SGCls} & \multicolumn{3}{c}{SGDet} \\
                & & zR@20 & zR@50 & zR@100
               & zR@20 & zR@50 & zR@100
               & zR@20 & zR@50 & zR@100 \\
                \hline
                \hline
               \multirow{5}{*}{\begin{sideways}VGG-16\end{sideways}}
                &
                IMP~\cite{xu2017scene}
                & -  & 14.5 & 17.2
                & -  & 2.5  & 3.2
                & -  & -    & 0.9 \\
                & Motifs~\cite{zellers2018neural}
                & -  & 6.5 	& 9.5
                & -  & 1.1  & 1.7
                & -  & -    & 0.3 \\
                & IMP++~\cite{knyazev2020graph}
                & -  & 18.4	& 21.5
                & -  & 3.4  & 4.2
                & -  & -    & 0.8 \\
                & GRAPHN~\cite{knyazev2021generative}
                & -  & \textcolor{myblue}{\textbf{19.5}}	& \textcolor{myblue}{\textbf{22.4}}
                & -  & \textcolor{myblue}{\textbf{3.8}}  & \textcolor{myblue}{\textbf{4.5}} 
                & -  & - & \textcolor{myblue}{\textbf{1.1}} \\
                & \cellcolor{mygray-bg}{\textbf{T-CAR (ours)}}
                & \cellcolor{mygray-bg}{\textcolor{red}{\textbf{23.1}}} 
                & \cellcolor{mygray-bg}{\textcolor{red}{\textbf{29.6}}} 
                & \cellcolor{mygray-bg}{\textcolor{red}{\textbf{32.8}}}
                & \cellcolor{mygray-bg}{\textcolor{red}{\textbf{6.4}}} 
                & \cellcolor{mygray-bg}{\textcolor{red}{\textbf{7.6}}} 
                & \cellcolor{mygray-bg}{\textcolor{red}{\textbf{8.7}}}
                & \cellcolor{mygray-bg}{\textcolor{red}{\textbf{2.4}}} 
                & \cellcolor{mygray-bg}{\textcolor{red}{\textbf{3.4}}} 
                & \cellcolor{mygray-bg}{\textcolor{red}{\textbf{4.2}}} \\
                \hline
                \multirow{12}{*}{\begin{sideways}X-101-FPN\end{sideways}}
                & IMP~\cite{xu2017scene}
                & 12.3 & 17.5 & 19.9
                & 1.2 & 1.9 & 2.2
                & 0.1 & 0.5 & 0.9 \\
                & VTransE~\cite{zhang2017visual}
                & 7.0 & 11.7 & 15.0
                & 1.1 & 1.9 & 2.3
                & 0.5 & 0.9 & 1.6 \\
                & Motifs~\cite{zellers2018neural}
                & 11.8 & 17.7 & 20.5
                & 2.6 & 4.1 & 5.0
                & 0.9 & 1.9 & 2.7 \\
                & Motifs+Freq~\cite{zellers2018neural}
                & 0.1 & 2.8 & 4.8
                & 0.4 & 0.7 & 1.1
                & 0.0 & 0.0 & 0.2 \\
                & IMP++~\cite{knyazev2020graph}
                & \textcolor{myblue}{\textbf{12.9}} & \textcolor{myblue}{\textbf{19.2}} & \textcolor{myblue}{\textbf{22.4}}
                & 2.6 & 4.2 & 5.0
                & 0.1 & 0.4 & 0.9 \\
                & TDE~\cite{tang2020unbiased}
                & 7.7 & 12.5 & 16.4
                & 1.6 & 2.6 & 3.5
                & 1.1 & 2.0 & 2.6 \\
                & UVTransE~\cite{hung2021contextual}
                & 10.7 & 16.5 & 18.9
                & 2.2 & 3.3 & 3.9
                & 0.6 & 1.2 & 2.1 \\
                & EBM~\cite{suhail2021energy}
                & 11.3 & 16.8 & 20.0
                & 3.4
                & 5.3 & \textcolor{myblue}{\textbf{6.2}}
                & 1.0 & 2.0 & 3.0 \\
                & BGNN~\cite{li2021bipartite}
                & 1.5 & 3.5 & 5.2
                & 0.9 & 1.7 & 2.2
                & 0.1 & 0.1 & 0.3 \\
                & SSR(Base)~\cite{teng2022structured}
                & - & - & -
                & - & - & -
                & 1.6 & 2.6 & 3.6 \\
                & SSR(Large)~\cite{teng2022structured}
                & - & - & -
                & - & - & -
                & 1.8
                & 2.8
                & \textcolor{myblue}{\textbf{4.2}} \\
                & NARE\dag~\cite{goel2022not}
                & 9.1 & 13.5 & -
                & \textcolor{myblue}{\textbf{4.3}} 
                & \textcolor{myblue}{\textbf{6.2}} & -
                & \textcolor{myblue}{\textbf{2.2}} 
                & \textcolor{myblue}{\textbf{3.3}} & - \\
                & \cellcolor{mygray-bg}{\textbf{T-CAR (ours)}}
                & \cellcolor{mygray-bg}{\textcolor{red}{\textbf{24.5}}}  
                &  \cellcolor{mygray-bg}{\textcolor{red}{\textbf{31.9}}} 
                & \cellcolor{mygray-bg}{{\textcolor{red}{\textbf{34.9}}}}
                & \cellcolor{mygray-bg}\textcolor{red}{\textbf{6.9}}
                & \cellcolor{mygray-bg}\textcolor{red}{\textbf{9.3}} 
                & \cellcolor{mygray-bg}{\textcolor{red}{\textbf{10.6}}}
                & \cellcolor{mygray-bg}\textcolor{red}{\textbf{3.2}} 	
                &  \cellcolor{mygray-bg}\textcolor{red}{\textbf{4.7}} 
                & \cellcolor{mygray-bg}{\textcolor{red}{\textbf{6.0}}} \\
                \hline
            \end{tabular}
        }
    \label{tab:compare_with_sota}
\end{table}

\begin{figure}[!t]
    \centering
    \includegraphics[width = 1.0 \columnwidth]{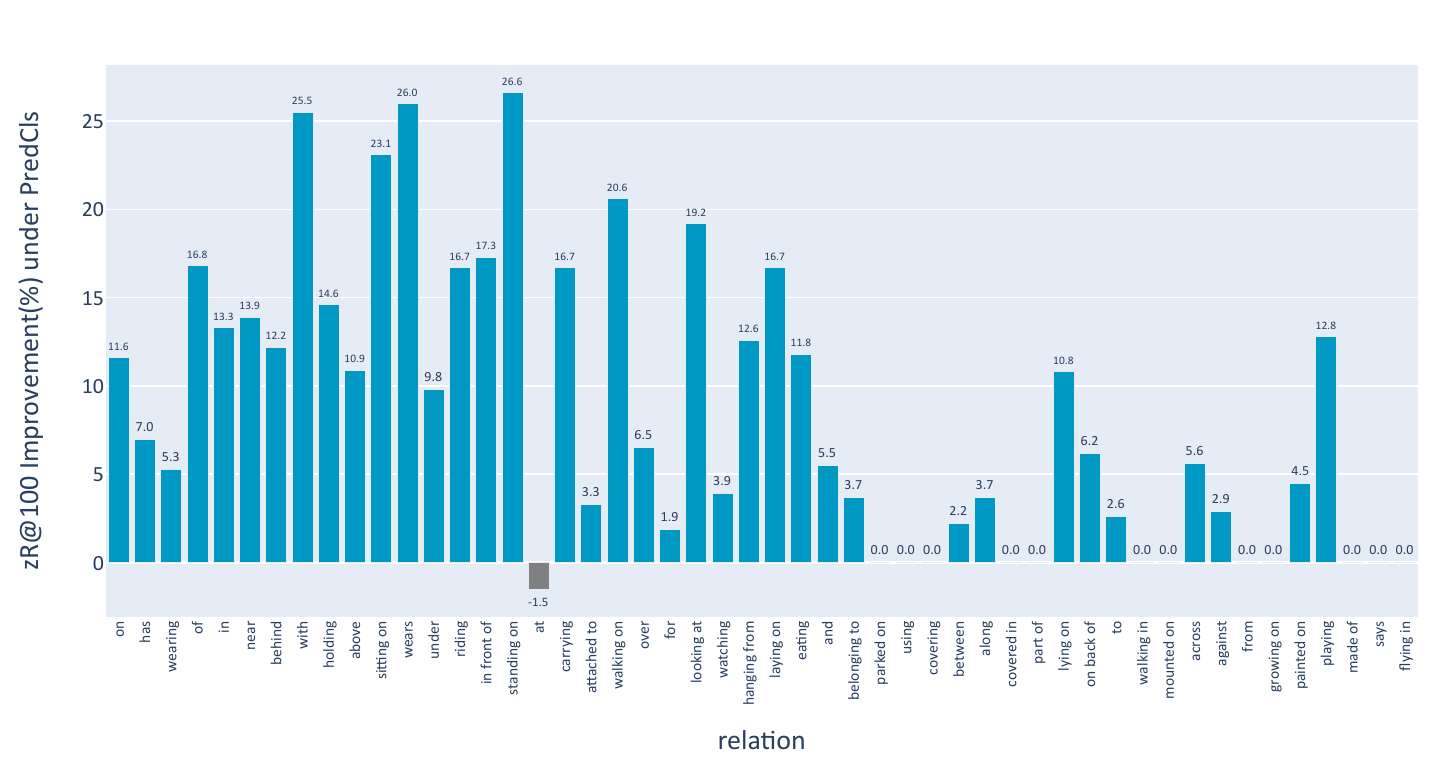}
    \caption{
Absolute zR@100 improvement in PredCls task by T-CAR compared to IMP++ \cite{knyazev2020graph} with ResNeXt-101-FPN backbone. The predicate categories are sorted according to their frequency.
    }
    \label{fig:improve-bar-predcls}
\end{figure}

\begin{figure}[!t]
    \centering
    \includegraphics[width = 1.0 \columnwidth]{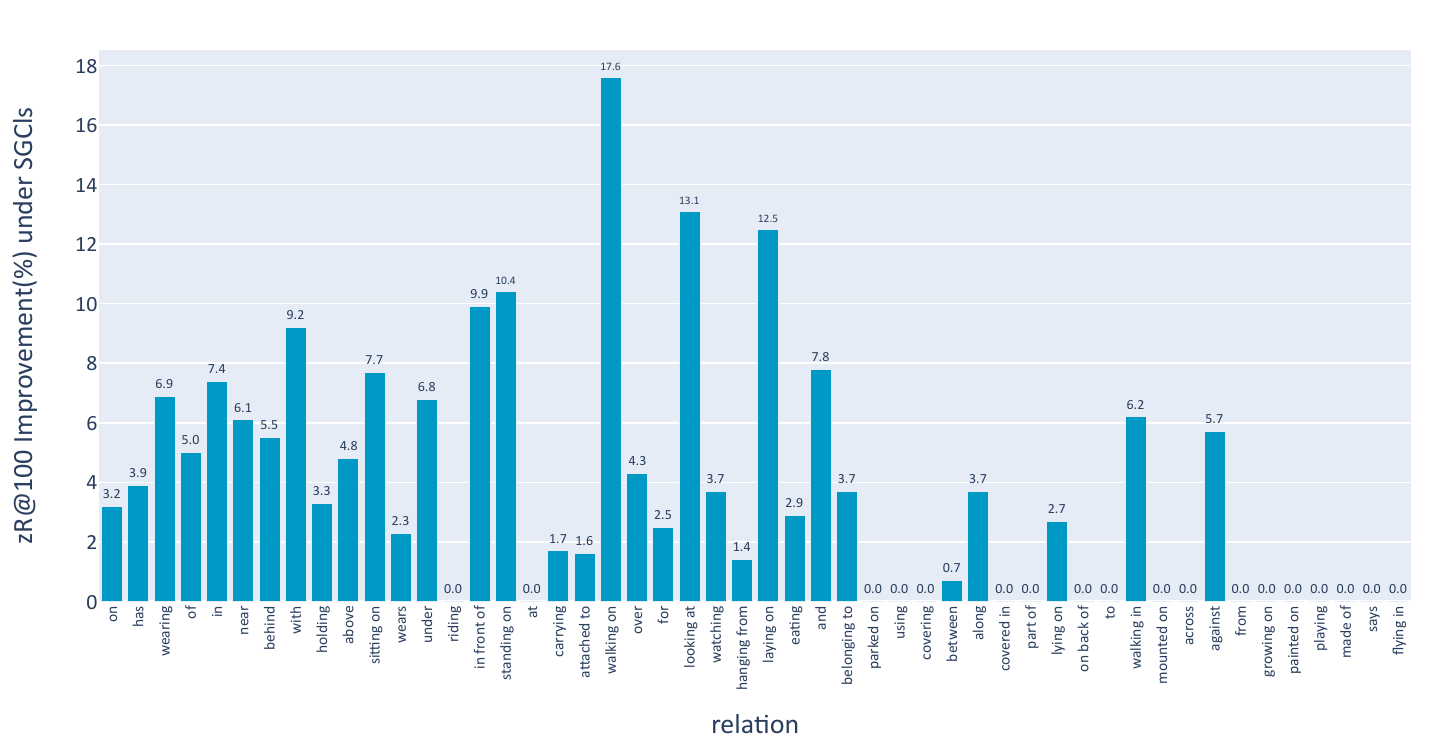}
    \caption{
Absolute zR@100 improvement in SGCls task by T-CAR compared to IMP++ \cite{knyazev2020graph} with ResNeXt-101-FPN backbone. The predicate categories are sorted according to their frequency.
    }
    \label{fig:improve-bar-sgcls}
\end{figure}

\subsection{Comparisons with State-of-the-Art Methods}
We compare several state-of-the-art methods on the Visual Genome dataset to demonstrate the effectiveness of our approach. IMP \cite{xu2017scene}, VTransE \cite{zhang2017visual}, and Motifs \cite{zellers2018neural} focus on all triplet prediction, so their performance on seen triplets is much better than on unseen triplets. IMP++ \cite{knyazev2020graph}, TDE \cite{tang2020unbiased}, UVTransE \cite{hung2021contextual}, EBM \cite{suhail2021energy}, GRAPHN \cite{knyazev2021generative}, SSR \cite{teng2022structured}, and NARE \cite{goel2022not} aim to generate unseen triplets. BGNN \cite{li2021bipartite} is a state-of-the-art method focusing on unbiased predicate prediction. 

\begin{table*}[tbp]
    \centering
    \caption{Performance comparison results of our T-CAR method with state-of-the-art SGG models on three SGG tasks without graph constraint. ``B" denotes the backbone of object detector (Faster R-CNN~\cite{ren2015faster}) used in each SGG model. ``-'' represents that the result is not mentioned in the original paper or the method is unavailable in that configuration.}
        \resizebox{14cm}{!}{
            \begin{tabular}{c| r | c c c | c  c  c | c c c}
                \hline
                \multirow{2}{*}{B} &
                \multirow{2}{*}{Models} & \multicolumn{3}{c|}{PredCls} & \multicolumn{3}{c|}{SGCls} & \multicolumn{3}{c}{SGDet} \\
                & & zR@20 & zR@50 & zR@100
               & zR@20 & zR@50 & zR@100
               & zR@20 & zR@50 & zR@100 \\
                \hline
                \hline
                \multirow{11}{*}{\begin{sideways}X-101-FPN\end{sideways}}
                & IMP~\cite{xu2017scene}
                & 14.4 & 27.5 & 38.9
                & 1.5 & 3.8 & 6.8
                & 0.1 & 0.5 & 1.1 \\
                & VTransE~\cite{zhang2017visual}
                & 8.2 & 17.9 & 28.6
                & 1.6 & 3.9 & 7.0
                & 0.9 & 1.7 & 3.2 \\
                & Motif~\cite{zellers2018neural}
                & 14.3 & 27.4 & 39.7
                & 3.1 & 6.9 & 11.2
                & 1.1 & 2.7 & 4.5 \\
                & IMP++~\cite{knyazev2020graph}
                & 14.6 & 27.3 & 39.0
                & 3.1 & 7.5 & 11.4
                & 0.1 & 0.4 & 0.9 \\
                & TDE~\cite{tang2020unbiased}
                & 8.9 & 17.4 & 26.7
                & 1.3 & 4.2 & 8.3
                & 0.0 & 0.1 & 0.4 \\
                & EBM~\cite{suhail2021energy}
                & 13.8 & 26.2 & 38.3
                & 4.1 & 8.5 & 13.9
                & 1.3 & 2.7 & 4.4 \\
                & UVTransE~\cite{hung2021contextual}
                & 12.7 & 25.4 & 37.3
                & 2.7 & 5.9 & 9.8
                & 0.9 & 2.3 & 3.9 \\
                & BGNN~\cite{li2021bipartite}
                & 2.2 & 7.5 & 15.6
                & 1.3 & 3.9 & 6.8
                & 0.1 & 0.2 & 0.5 \\
                & SSR(Base)~\cite{teng2022structured}
                & - & - & -
                & - & - & -
                & 2.0 & 4.1 & 6.0 \\
                & SSR(Large)~\cite{teng2022structured}
                & - & - & -
                & - & - & -
                & 2.1 & 4.0 & 5.9 \\
                & NARE~\cite{goel2022not}
                & - & - & -
                & - & - & -
                & - & - & - \\
                & \cellcolor{mygray-bg}{\textbf{T-CAR (ours)}}
                & \cellcolor{mygray-bg}{\textbf{27.4}}
                &  \cellcolor{mygray-bg}{\textbf{39.8}}
                & \cellcolor{mygray-bg}{{\textbf{50.8}}}
                & \cellcolor{mygray-bg}\textbf{7.8}
                & \cellcolor{mygray-bg}\textbf{12.4}
                & \cellcolor{mygray-bg}{\textbf{16.7}}
                & \cellcolor{mygray-bg}\textbf{3.6}
                &  \cellcolor{mygray-bg}\textbf{6.2}
                & \cellcolor{mygray-bg}{\textbf{8.7}} \\
                \hline
            \end{tabular}
        }
    \label{tab:compare_wo_gc}
\end{table*}

\begin{table*}[tbp]
    \centering
    \caption{Performance comparison results of our T-CAR method with state-of-the-art SGG models on three SGG tasks with graph constraint. ``B" denotes the backbone of object detector (Faster R-CNN~\cite{ren2015faster}) used in each SGG model. \dag \, denotes that the results are obtained with an unknown evaluation protocol, and thus, may not be directly comparable. ``-'' represents that the result is not mentioned in the original paper or the method is unavailable in that configuration.}
        \resizebox{14cm}{!}{
            \begin{tabular}{c| r |c c | c  c|c c}
                \hline
                \multirow{2}{*}{B} &
                \multirow{2}{*}{Models} & \multicolumn{2}{c|}{PredCls} & \multicolumn{2}{c|}{SGCls} & \multicolumn{2}{c}{SGGen} \\
                & 
                & zR@50/100 & R@50/100
                & zR@50/100 & R@50/100
                & zR@50/100 & R@50/100 \\
                \hline
                \hline
                \multirow{11}{*}{\begin{sideways}X-101-FPN\end{sideways}}
                & IMP~\cite{xu2017scene}
                & 17.5 / 19.9 & 61.3 / 63.3
                & 1.9 / 2.2 & 61.3 / 35.3
                & 0.5 / 0.9 & 25.7 / 31.2 \\
                & VTransE~\cite{zhang2017visual}
                & 11.7 / 15.0 & 58.2 / 62.6
                & 1.9 / 2.3 & 33.4 / 35.6
                & 0.9 / 1.6 & 27.2 / 31.5 \\
                & Motif~\cite{zellers2018neural}
                & 17.7 / 20.5 & \textbf{65.0} / \textbf{66.9}
                & 4.1 / 5.0 & 39.1 / 39.9
                & 1.9 / 2.7 & \textbf{32.6} / \textbf{37.0} \\
                & IMP++~\cite{knyazev2020graph}
                & 19.2 / 22.4 & 62.1 / 64.3
                & 4.3 / 5.1 & 40.0 / 40.8
                & 0.4 / 0.9 & 21.1 / 27.4 \\
                & TDE~\cite{tang2020unbiased}
                & 12.5 / 16.4 & 45.7 / 51.1
                & 2.6 / 3.5 & 28.0 / 30.5
                & 2.0 / 2.6 & 16.7 / 20.3 \\
                & EBM~\cite{suhail2021energy}
                & 16.8 / 20.0 & 64.5 / 66.5
                & 5.3 / 6.2 & \textbf{43.5} / \textbf{44.7}
                & 2.0 / 3.0 & 30.3 / 34.6 \\
                & UVTransE~\cite{hung2021contextual}
                & 16.5 / 18.9 & 64.7 / 66.4
                & 3.3 / 3.9 & 37.9 / 38.8
                & 1.2 / 2.1 & 31.9 / 36.1 \\
                & BGNN~\cite{li2021bipartite}
                & 3.5 / 5.2 & 58.1 / 60.9
                & 1.7 / 2.2 & 36.2 / 37.4
                & 0.1 / 0.3 & 25.9 / 31.1 \\
                & SSR(Base)~\cite{teng2022structured}
                & - / - & - / -
                & - / - & - / -
                & 2.6 / 3.6 & 23.3 / 26.5 \\
                & SSR(Large)~\cite{teng2022structured}
                & - / - & - / -
                & - / - & - / -
                & 2.8 / 4.2 & 23.7 / 27.3 \\
                & NARE\dag~\cite{goel2022not}
                & 13.5 / - \, \, \, \ & 47.6 / 52.0
                & 6.2 / - \, \ \ & 32.8 / 35.8
                & 3.3 / - \, \ \ & 19.0 / 21.0 \\
                & \cellcolor{mygray-bg}{\textbf{T-CAR (ours)}}
                & \cellcolor{mygray-bg}\textbf{31.9} / \cellcolor{mygray-bg}\textbf{34.9}
                & \cellcolor{mygray-bg}{60.0} / \cellcolor{mygray-bg}{63.0}
                & \cellcolor{mygray-bg}\textbf{9.3} / \cellcolor{mygray-bg}\textbf{10.6}
                & \cellcolor{mygray-bg}{40.4} / \cellcolor{mygray-bg}{42.0} 
                & \cellcolor{mygray-bg}\textbf{4.7} / \cellcolor{mygray-bg}\textbf{6.0}
                & \cellcolor{mygray-bg}{28.5} / \cellcolor{mygray-bg}{32.9} \\
                \hline
            \end{tabular}
        }
    \label{tab:compare_gc}
\end{table*}

We compare BGNN with our method to demonstrate the difference between unbiased scene graph generation and zero-shot scene graph generation, \ie \ methods that concentrate solely on unbiased scene graph generation do not work well on zero-shot scene graph generation task. It is worth noting that GRAPHN \cite{knyazev2021generative} generates visual feature maps to augment the input scene graphs and improve the generalization of its scene graph model on unseen triplets. Here we only compare GRAPHN \cite{knyazev2021generative} with VGG-16 backbone since we argue that it is challenging to extend the generative model of GRAPHN from one scale to multiple scales with Feature Pyramid Network (FPN). Due to the limitation of SSR, its comparison is only possible under the SGDet task.

\subsubsection{Comparison with Graph Constraint}
Tab. \ref{tab:compare_with_sota} shows the comparison results of our model on the Visual Genome dataset with graph constraint.
We have the following observations: 

1) Our T-CAR model consistently and significantly outperforms all state-of-the-art methods on all three tasks with two backbones, which achieves a large margin of improvements by 12.5\%, 4.2\%, and 1.8\% on zR@100 for PredCls, SGCls, and SGDet, respectively.
Compared with the predicate granularity-based debiasing method \cite{tang2020unbiased} and unlabeled sample mining method \cite{goel2022not}, our triplet granularity-based T-CAR significantly improves the compositional generalization ability, which indicates that it is better to solve the zero-shot SGG problem at the triplet granularity.
Our method surpasses the approach Structured Sparse R-CNN (SSR) \cite{teng2022structured} that directly predicts triplets and GAN-based model GRAPHN \cite{knyazev2021generative}, demonstrating the effectiveness of excavating unseen triplets in the training set, which greatly reduces the seen triplet bias.

2) Compared with methods aiming to address the problem of imbalanced predicates, \ie \ BGNN \cite{li2021bipartite}, we witness that its performance is much lower than the SGG baselines, namely IMP \cite{xu2017scene}, Motifs \cite{zellers2018neural}, and UVTransE \cite{hung2021contextual}, indicating that the problem of zero-shot SGG is not equivalent to the issue of imbalanced predicates.

3) Deleting Frequent Bias on Motifs severely boosts its performance in zero-shot Recall. It demonstrates that the Frequent Bias module forces the models to bias toward seen triplets and decays their generalization capabilities.

\begin{table}
    \centering
    \caption{Ablation studies on each component of T-CAR. We use the same object detection backbone as in \cite{li2021bipartite}.}
        \scalebox{0.9}{
            \begin{tabular}{c c c| c c c | c c c}
                \hline
                \multicolumn{3}{c|}{Module} & \multicolumn{3}{c|}{SGCls} & \multicolumn{3}{c}{PredCls}  \\
                CEN & TCL & USRL
                & zR@20 & zR@50 & zR@100
                & zR@20 & zR@50 & zR@100 \\
                \hline
                \hline
                \XSolidBrush  & \XSolidBrush & \XSolidBrush
                & 1.7  & 3.1  & 3.9
                & 8.0  & 13.2 & 16.6 \\
                \Checkmark  & \XSolidBrush & \XSolidBrush
                & 3.4 & 5.3  & 6.4
                & 14.7 & 21.0  & 24.3 \\
                \XSolidBrush  & \Checkmark & \XSolidBrush
                & 3.9  & 5.9  & 7.3
                & 16.5  & 22.6  & 26.7 \\
                \Checkmark  & \Checkmark & \XSolidBrush
                & 6.3  & 8.6  & 9.8
                & 24.4  & 31.0 & 34.1 \\
                \XSolidBrush  & \Checkmark & \Checkmark
                & 4.1  & 6.3  & 7.9
                & 16.1  & 23.1  & 27.6 \\
                \hline
                \Checkmark  & \Checkmark & \Checkmark
                & \textbf{6.9}  & \textbf{9.3}  & \textbf{10.6}
                & \textbf{24.5}  & \textbf{31.9} & \textbf{34.9} \\
                \hline
            \end{tabular}
        }
    \label{tab:ablation_all}
\end{table}

\begin{table}[tbp]
    \centering
    \caption{Ablation studies on the margin in TCL.}
        \scalebox{0.9}{
            \begin{tabular}{c c| c c c | c c c}
                \hline
                \multicolumn{2}{c|}{Module} & \multicolumn{3}{c|}{SGCls} & \multicolumn{3}{c}{PredCls}  \\
                MU & MCE
                & zR@20 & zR@50 & zR@100
                & zR@20 & zR@50 & zR@100 \\
                \hline
                \hline
                \XSolidBrush  & \XSolidBrush
                & 6.1  & 8.8  & 10.0
                & 23.0  & 30.3  & 34.0 \\
                \Checkmark  & \XSolidBrush
                & 6.3  & 8.8  & 10.0
                & 22.7  & 30.5  & 34.2 \\
                \XSolidBrush  & \Checkmark
                & 6.5  & 9.0  & 10.1
                & 24.5  & 31.6  & 34.8 \\
                \hline
                \Checkmark  & \Checkmark
                & \textbf{6.9}  & \textbf{9.3}  & \textbf{10.6}
                & \textbf{24.5}  & \textbf{31.9} & \textbf{34.9} \\
                \hline
            \end{tabular}
        }
    \label{tab:ablation_cal}
\end{table}

\subsubsection{Improvements on Each Predicate Category}
As shown in Fig. \ref{fig:improve-bar-predcls}, we further investigate the improvement of our method over the IMP++ \cite{knyazev2020graph} on each predicate category to show its performance on different predicates. We find that T-CAR significantly improves the performance in head, body, and tail predicate categories. It is fundamentally different from approaches solving imbalanced predicates that typically boost the performance of tail classes and weaken the performance of the head ones.

\subsubsection{Comparison without Graph Constraint}
The setting of \textit{without graph constraint} is proposed by Zellers \cite{zellers2018neural}. It allows the output scene graph to contain multiple edges between the subject and object, as opposed to \textit{graph constraint}.
Higher recall performance can usually be obtained without the graph constraint since the model is allowed to have multiple guesses for challenging relations.
To extensively analyze the compositional generalization ability of our T-CAR method, we also report the comparison results without graph constraints.
As shown in Tab. \ref{tab:compare_wo_gc}, with multiple guesses for challenging relations, our T-CAR method consistently outperforms state-of-the-art methods on all three tasks without graph constraint, demonstrating the superior compositional generalization ability of our method. 

\subsubsection{Comparison with Relationship Recall}
The full results of Relationship Recall with graph constraint, including both conventional Recall@K and the adopted Zero-Shot Recall@K, are shown in Tab. \ref{tab:compare_gc}. 

IMP \cite{xu2017scene}, VTransE \cite{zhang2017visual}, and Motifs \cite{zellers2018neural} focus on all triplet predictions and achieve better results on Recall than other methods.
We can observe that our method achieves state-of-the-art performance on unseen triplets with acceptable performance degradation on seen samples.
Compared with the recently proposed state-of-the-art approaches, \ie \ SSR \cite{teng2022structured} and NARE \cite{goel2022not}, our T-CAR model not only owns better compositional generalization ability but also significantly outperforms them on the seen samples.

\begin{table}[tbp]
    \centering
    \caption{Ablation studies on the reduction rate.}
        \scalebox{0.9}{
            \begin{tabular}{c| c c c c c c c c}
                \hline
                 ~ & \multicolumn{8}{c}{Reduction Rate} \\
                 ~ & 0\% & 50\% & 60\% & 70\% & 80\% & 85\% & 90\% & 100\% \\
                \hline
                \hline
                 zR@20 & 6.3 & 6.4  & 6.6 & 6.5 & 6.7 & \textbf{6.9} & 6.6 & 3.3 \\
                 zR@50 & 8.6 & 8.7  & 8.8 & 8.8 & 9.2 & \textbf{9.3} & 8.7 & 5.0 \\ 
                 zR@100 & 9.8 & 10.0  & 10.1 & 10.0 & 10.5 & \textbf{10.6} & 10.3 & 6.1 \\
                \hline
            \end{tabular}
        }
    \label{tab:ablation_reduce_rate}
\end{table}

\begin{table}[tbp]
    \centering
    \caption{Ablation studies on the unseen space reduction method.}
        \scalebox{0.9}{
            \begin{tabular}{c | c c c c}
                \hline
                Method & AUC & Recall & Precision & F1-score  \\
                \hline
                \hline
                Random
                & 49.3  & 51.9
                & 0.5  & 1.0 \\
                BiLSTM
                & 91.0  & 72.6
                & 3.3  & 6.1 \\
                Ours
                & \textbf{93.0}  & \textbf{73.8}
                & \textbf{4.2}  & \textbf{7.9} \\
                \hline
            \end{tabular}
        }
    \label{tab:ablation_reduce}
\end{table}

\subsection{Ablation Studies}

\subsubsection{Model Components.}
We conduct an ablation study to evaluate the importance of each component in our T-CAR, \ie \ Contextual Encoding Network (CEN), Triplet Calibration Loss (TCL), and Unseen Space Reduce Loss (USRL). The results are shown in Tab. \ref{tab:ablation_all}. 
Specifically, we remove all modules in T-CAR and use a baseline without explicit relation feature refinement. This baseline predicts predicates with visual features of pairwise entities, and its performance is lower than any other variants of T-CAR. Then we add these proposed components to the baseline method. 

As shown in Tab. \ref{tab:ablation_all}, all modules promote performance, and the best performance is achieved when all modules are involved.
We observe that the TCL improves the CEN and achieves 7.9 and 32.0 on SGCls and PredCls tasks, demonstrating that reducing the seen triplet bias and relaxing constraints on potentially unseen samples can facilitate the zero-shot SGG. 
Compared with CEN+TCL, we witness an obvious performance gain with USRL. 
Note that the USRL is designed to reduce the search space of unseen triplets, so its relative boost on SGCls is greater than that of PredCls. Moreover, applying CEN alone is able to achieve similar performance to state-of-the-art methods \cite{knyazev2020graph,suhail2021energy}. It verifies that more robust position encoding can alleviate the seen triplet bias, leading to more accurate predictions.

\subsubsection{Margin of Calibration.}
We also evaluate the effectiveness of weighted seen triplet margins on TCL and report the results in Tab. \ref{tab:ablation_cal}. 
We first set the margins of seen triplets in TCL equal to 1 as the baseline. 
Then margin constraints on seen triplets and unseen triplets calibration are added, named MU and MCE, respectively. 
The results show that changing the margins on seen samples by their occurrence frequency in both cross-entropy and unseen sample mining losses can improve performance. Frequent triplets in scene graph do affect the generalization of unseen samples.
Best performance is achieved when both MU and MCE are engaged.

\begin{table}[tbp]
    \centering
    \caption{Ablation studies on the initialization of features of CEN and T-CAR. The w/o P and w L denote initialize features without relative positional encoding and with the linguistic feature, respectively.}
        \scalebox{0.9}{
            \begin{tabular}{c| c c c | c c c}
                \hline
                \multirow{2}{*}{Model} & \multicolumn{3}{c|}{SGCls} & \multicolumn{3}{c}{PredCls}  \\
                ~ 
                & zR@20 & zR@50 & zR@100
                & zR@20 & zR@50 & zR@100 \\
                \hline
                \hline
                CEN w/o P
                & 3.2 & 5.2  & 6.3
                & 14.0 & 20.8 & 24.1 \\
                CEN w L
                & 3.2 & 5.2  & 6.4
                & 13.0 & 19.6  & 23.1 \\
                CEN
                & 3.4 & 5.3  & 6.4
                & 14.7 & 21.0  & 24.3 \\
                \hline
                T-CAR w/o P
                & 6.0 & 8.5  & 9.9
                & \textbf{24.8} & 31.4  & 34.7 \\
                T-CAR w L
                & 6.5 & 8.8  & 10.0
                & 24.4 & 31.5  & 34.8 \\
                T-CAR
                & \textbf{6.9}  & \textbf{9.3}  & \textbf{10.6}
                & 24.5 & \textbf{31.9} & \textbf{34.9} \\
                \hline
            \end{tabular}
        }
    \label{tab:ablation_pos}
\end{table}

\begin{table}[tbp]
    \centering
    \caption{Ablation studies on the hyper-parameter $\lambda$ in TCL.}
        \scalebox{0.9}{
            \begin{tabular}{c| c c c | c c c}
                \hline
                \multirow{2}{*}{$\lambda$} & \multicolumn{3}{c|}{SGCls} & \multicolumn{3}{c}{PredCls}  \\
                ~
                & zR@20 & zR@50 & zR@100
                & zR@20 & zR@50 & zR@100 \\
                \hline
                \hline
                0.001
                & 6.3  & 8.5  & 9.8
                & 23.9  & 31.0  & 34.3 \\
                0.01
                & \textbf{6.9}  & \textbf{9.3}  & \textbf{10.6}
                & \textbf{24.5}  & \textbf{31.9} & \textbf{34.9} \\
                0.1
                & 5.3  & 7.4  & 9.0
                & 23.4  & 29.3  & 33.3 \\
                1.0
                & 2.2  & 4.2  & 5.8
                & 10.4  & 16.5  & 22.3 \\
                \hline
            \end{tabular}
        }
    \label{tab:ablation_lambda}
\end{table}

\subsubsection{Unseen Space Reduction Loss.}
In Tab. \ref{tab:ablation_reduce}, we examine the effect of the interchangeability module in USRL on the reduction of unseen triplet space. Unseen triplets in the VG test set are treated as test samples. We apply the ranking metric AUC and commonly used evaluation metrics for binary classification, \ie \ Recall, Precision, and F1, to evaluate model performance. The precision is very low since the entire space is extremely large, and the unseen triplets in the VG test set are not equivalent to all positive unseen samples.

``Random'' in Tab. \ref{tab:ablation_reduce} implies that we apply a randomly initialized MLP to assign the triplets of the unseen space as reasonable or unreasonable without the knowledge of seen realistic triplets. We perform random strategy five times and report their average results.
In addition, we also apply a two-layer BiLSTM on triplets as the baseline method. Compared with the ``random'', it is observed that BiLSTM can learn from seen triplets and judge the reasonableness of unknown triplets. BiLSTM \cite{hochreiter1997long} is able to rank the reasonableness of triplets and achieves good performance.
Compared with BiLSTM, our unseen space reduction improves the performance of all metrics, especially in AUC. It indicates that our method performs better in ranking the plausibility of triplets. We own this advantage to the alternative mining of triplets in subject, predicate, and object.

In Tab. \ref{tab:ablation_reduce_rate}, we also explore the impact of reduction rate in USRL for SGCls task. 
Unseen triplet space gradually decreases as the reduction rate increases, and the model shifts its attention to excavating reasonable triplets. But when the reduction rate is larger than a certain degree, it will inevitably misclassify some of the unseen and reasonable triplets. Thus the performance of T-CAR behaves as an increase followed by a decrease, and the best results are achieved when the reduction rate is set as 85\%.

\subsubsection{Ablation Study on Hyper-Parameter in TCL}

$\lambda$ in Eq. \ref{eq:all-loss} controls the attention of the model to unlabeled and unseen samples, which is important for unseen triplets excavation.
We study the effect of this hyper-parameter on SGCls and PredCls tasks with the ResNeXt-101-FPN backbone. 
As shown in Tab. \ref{tab:ablation_lambda},
our model gradually increases its attention to the unseen samples as the $\lambda$ value increases.
When the $\lambda$ reaches a certain value, the attention to unseen samples affects the learning on seen samples, which leads to the weakened ability to represent diverse triplets and decreased performance on unseen samples. The best performance is achieved when $\lambda$ equals $0.01$.

\subsubsection{Ablation Study on Initialization of Features}
We verify the effectiveness of initialized features in CEN and T-CAR on compositional generalization ability with the ResNeXt-101-FPN backbone.
As shown in Tab. \ref{tab:ablation_pos}, both the T-CAR model and CEN have better compositional generalization ability without linguistic features and with relative positional encoding.
CEN with linguistic features on the PredCls task encounters a severe drop in unseen performance, indicating that the subject and object priors introduced by linguistic features do affect the compositional generalization ability.

\begin{figure}[!t]
    \centering
    \includegraphics[width = 1.0 \columnwidth]{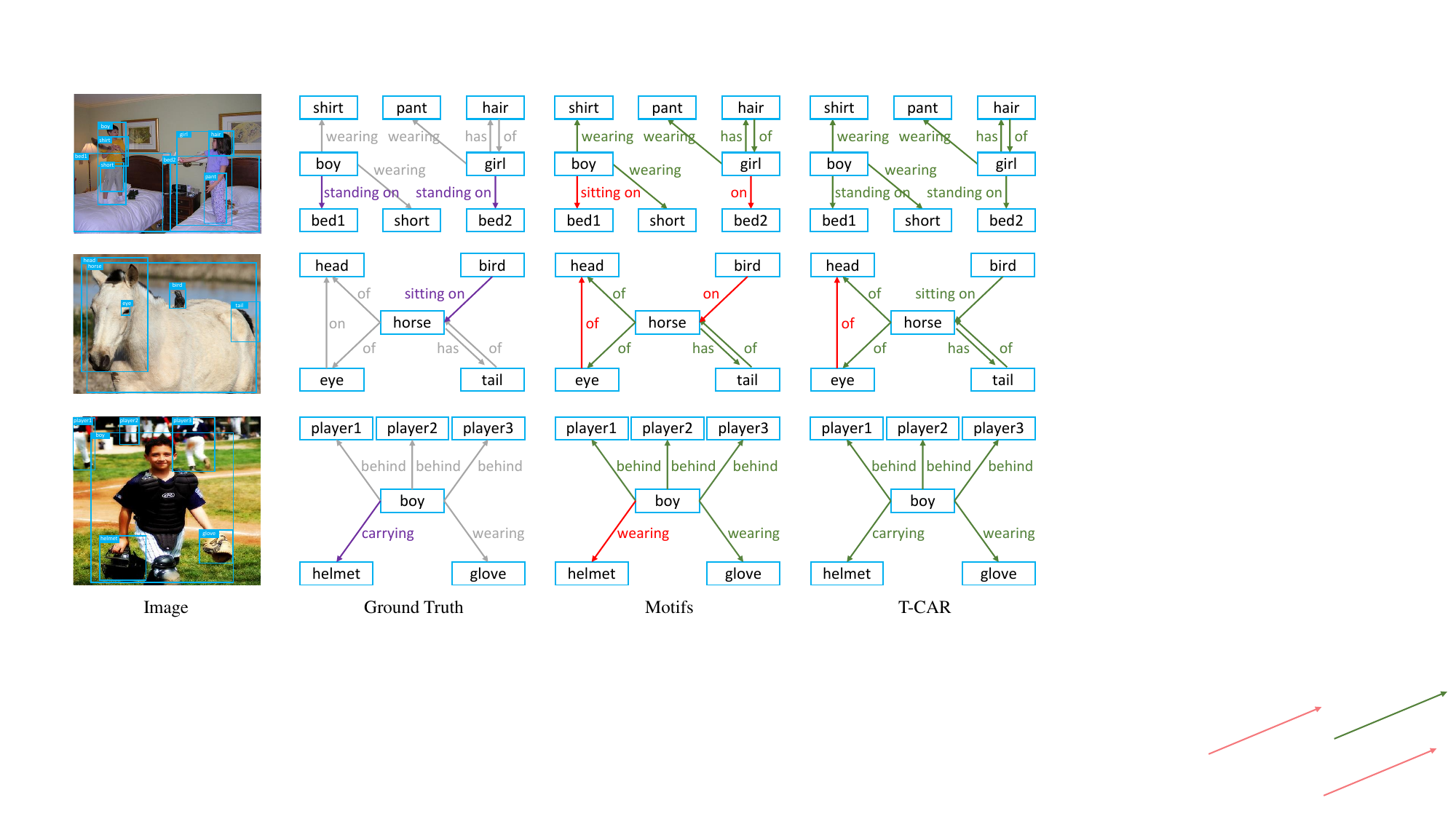}
    \caption{
Qualitative comparisons between our T-CAR and Motifs \cite{zellers2018neural} in the PredCls setting. The purple color indicates the unseen triplets in test images. The green color denotes the correctly classified triplets, and the red suggests the misclassified triplets. Best viewed in color.
    }
    \label{fig:qualitative}
\end{figure}

\begin{table}[tbp]
    \centering
    \caption{Analysis on the knowledge transfer. We add the recently proposed distribution knowledge transfer loss (D-TRS) in TN-ZSTAD \cite{Zhang2023TN} in our T-CAR, which further improves the performance of our model.}
        \scalebox{0.9}{
            \begin{tabular}{c| c c c | c c c}
                \hline
                \multirow{2}{*}{Model} & \multicolumn{3}{c|}{SGCls} & \multicolumn{3}{c}{PredCls}  \\
                ~ 
                & zR@20 & zR@50 & zR@100
                & zR@20 & zR@50 & zR@100 \\
                \hline
                \hline
                T-CAR 
                & 6.9  & \textbf{9.3}  & \textbf{10.6}
                & 24.5 & \textbf{31.9} & 34.9 \\
                T-CAR + D-TRS \cite{Zhang2023TN}
                & \textbf{7.2}  & 9.2  & \textbf{10.6}
                & \textbf{25.3} & 31.6 & \textbf{35.1} \\
                \hline
            \end{tabular}
        }
    \label{tab:analsy-tn}
\end{table}

\subsection{Analysis on the Knowledge Transfer}

Our triplet calibration loss serves the same purpose as the distribution transfer loss (D-TRS) in TN-ZSTAD \cite{Zhang2023TN}, \ie, regularizing the model to transfer knowledge from seen to unseen. D-TRS takes the semantic similarity of labels to prevent the classifier from over-confidence toward seen classes and to match the predicted probability distribution of unseen classes. Such semantic similarity between triplet labels can guide the model to mine the unseen triplets that are most likely to be misclassified as background, which can also collaborate with our method.

We follow D-TRS and take the CLIP \cite{clip2021} text embedding to calculate the semantic similarity between triplet labels. For instance, given a triplet label <Subject, Predicate, Object>, we generate the corresponding description in the format of `A photo of a/an [Subject] [Predicate] a/an [Object]’. Then, we generate the text embedding for each triplet label through the pre-trained CLIP text encoder. Finally, D-TRS is applied to obtain the similarity between labels and collaborate with our method. As shown in Tab. \ref{tab:analsy-tn}, the performance of our T-CAR method is further improved by D-TRS. Note that D-TRS boosts our method for zR@20 more than zR@100. It indicates that D-TRS gives much confidence to the unseen triplets mislabeled as backgrounds through the similarity between triplet labels in training.

\subsection{Qualitative Evaluation} 
We visualize the scene graphs generated by our T-CAR and compare them with Motifs \cite{zellers2018neural} to show the importance of the zero-shot SGG. Results are shown in Fig. \ref{fig:qualitative}. As seen from the first row, Motifs tends to predict the frequently seen relationships. It predicts the predicate between ``girl’' and ``bed2’' as ``on'’ instead of the more accurate ``standing on’', and misclassifies the relationships of ``boy’' and ``bed1'’ as the frequent predicate ``sitting on’'. T-CAR alleviates the seen triplets bias from frequent compositions and generates the correct category. As seen in the second row, T-CAR is equally effective in predicting the anthropomorphic actions made by animals. Finally, in the third row of Fig. \ref{fig:qualitative}, it can be observed that it is easy to exclude the predicate ``wearing’' and consider ``carrying’' based on the relative spatial information between ``boy’' and ``helmet’' by T-CAR. To sum up, our T-CAR makes better predictions than Motifs. We owe this performance gain of T-CAR to the explicit modeling of relative spatial features that alleviates the seen triplet bias in contextual encoding network.

\begin{table*}[tbp]
    \centering
    \caption{Model size and speed comparisons for SGDet. ``N/A’' denotes that the result is not available due to the limited GPU memory.
    }
        \resizebox{14cm}{!}{
            \begin{tabular}{c| r | c  c  c  c  c | c }
                \hline
                \multirow{2}{*}{Backbone} & \multirow{2}{*}{Models} & \multicolumn{5}{c|}{Training} & Inference \\
                ~ & ~ & \#Params (MB) & Iterations & Batch Size & Times (h) & Sec/image & FPS  \\
                \hline
                \hline
                \multirow{12}{*}{X-101-FPN}
                & IMP~\cite{xu2017scene}
                & 1,176 & 24,000 & 12
                & 14.4 & 0.09 & 5.6 \\
                & VTransE~\cite{zhang2017visual}
                & 1,170 & 28,000 & 12
                & 18.3 & 0.10 & 5.6 \\
                & Motifs~\cite{zellers2018neural}
                & 1,405 & 28,000 & 12
                & 21.8 & 0.13 & 4.2 \\
                & IMP++~\cite{knyazev2020graph}
                & 1,176 & 28,000 & 12
                & 19.5 & 0.12 & 5.4 \\
                & TDE~\cite{tang2020unbiased}
                & 1,414 & 28,000 & 12
                & 21.4 & 0.13 & 3.1 \\
                & UVTransE~\cite{hung2021contextual}
                & 1,217 & 20,000 & 12
                & 13.3 & 0.08 & 5.3 \\
                & EBM~\cite{suhail2021energy}
                & 1,419 & 200,000 & 4
                & 362.0 & 1.56 & 2.2 \\
                & BGNN~\cite{li2021bipartite}
                & 1,304 & 100,000 & 6
                & 105.1 & 0.62 & 2.5 \\
                & SSR(Base) ~\cite{teng2022structured}
                & 1,045 & 320,000 & 2
                & 127.1 & 0.71 & 11.8 \\
                & SSR(Large) ~\cite{teng2022structured}
                & 1,049 & N/A & N/A
                & N/A
                & N/A
                & 3.4 \\
                & T-CAR (ours)
                & 1,268 & 16,000 & 14
                & 11.6 & 0.12 & 3.9 \\
                \hline
            \end{tabular}
        }
    \label{tab:times}
\end{table*}

\subsection{Model Size and Speed}

Experiments are also conducted to analyze the model size and speed. Though scene graphs are powerful, it is time-consuming to perform SGG on the large-scale dataset. We include several previous works and run their codes under the same settings to analyze the model efficiency. Our experiments are conducted on two NVIDIA GeForce GTX 2080 Ti GPUs and an Intel Xeon E5-2650 v4 CPU. We set the training batch size to the maximum under the GPU memory limit, and the inference batch size is fixed to 2. Training SSR (Large) encounters the out-of-memory error due to its query numbers (300 for SSR (base) and 800 for SSR (Large)). Therefore, we only report its inference speed and model size. The results are shown in Table \ref{tab:times}. It is worth noting that our method requires only 16,000 iterations to converge, which is faster than all the other compared models. Our method also has fast inference speed and moderate parameter size. Overall, T-CAR performs well in terms of efficiency and is feasible for applications on large-scale datasets.

\section{Conclusion}
This paper introduces a Triplet Calibration and Reduction framework for zero-shot scene graph generation. It consists of a contextual encoding network, a triplet calibration loss, and an unseen space reduction loss. The contextual encoding network is based upon an entity encoder and a relation encoder. It explicitly models the relative spatial features between subjects and objects to alleviate seen triplet bias. The triplet calibration loss regularizes the representation of diverse triplets and mines the unseen triplets that are incorrectly annotated as background. Unseen Space Reduction Loss is built based on the interchangeability between seen triplets to reduce unreasonable triplets in unseen space. We also propose a new test protocol to facilitate a fair comparison of zero-shot SGG methods. Besides, both qualitative and quantitative evaluations are conducted to verify the effectiveness of the proposed method, and the results show that our method significantly outperforms the state-of-the-art zero-shot SGG methods on zero-shot triplets. In the future, we will explore leveraging external knowledge of large-scale pre-trained vision-language models, \eg \ CLIP \cite{clip2021}, to filter unreasonable triplets in the unseen space.

\section*{Acknowledgement}
This work is sponsored by the National Key R\&D Program of China (2022ZD0161901) and the National Nature Science Foundation of China (No. 62276018, U20B2069).

\bibliographystyle{ACM-Reference-Format}
\bibliography{acmart}


\end{document}